\documentclass{article} 
\usepackage{arxiv}

\usepackage[utf8]{inputenc} 
\usepackage[T1]{fontenc}    
\usepackage{hyperref}       
\usepackage{url}            
\usepackage{booktabs}       
\usepackage{amsfonts}       
\usepackage{nicefrac}       
\usepackage{microtype}      
\usepackage{xcolor}         
\usepackage{multirow}
\usepackage{graphicx}
\usepackage{amsmath, amssymb, algorithm, algorithmic, wrapfig}

\newcommand{\TimeSeriesExam}{\texttt{TimeSeriesExam}}

\title{\TimeSeriesExam: A Time Series Understanding Exam}

\author{
  Yifu Cai \quad Arjun Choudhry\thanks{Equal contribution} \quad Mononito Goswami\footnotemark[1] \quad Artur Dubrawski \\
  Auton Lab, School of Computer Science,\\ Carnegie Mellon University\\
  Pittsburgh, PA 15213\\
  \texttt{\{yifuc, arjuncho, mgoswami, awd\}@cs.cmu.edu} \\
}

\begin{document}
\maketitle
\begin{abstract}
Large Language Models (LLMs) have recently demonstrated a remarkable ability to model time series data. These capabilities can be partly explained if LLMs understand basic time series concepts. However, our knowledge of what these models understand about time series data remains relatively limited. 
To address this gap, we introduce \TimeSeriesExam, a configurable and scalable multiple-choice question exam designed to assess LLMs across five core time series understanding categories: \textit{pattern recognition}, \textit{noise understanding}, \textit{similarity analysis}, \textit{anomaly detection}, and \textit{causality analysis}. \TimeSeriesExam ~comprises of over 700 questions, procedurally generated using 104 carefully curated templates and iteratively refined to balance difficulty and their ability to discriminate good from bad models.
We test 7 state-of-the-art LLMs on the \TimeSeriesExam ~and provide the first comprehensive evaluation of their time series understanding abilities. Our results suggest that closed-source models such as \texttt{GPT-4} and \texttt{Gemini} understand simple time series concepts significantly better than their open-source counterparts, while all models struggle with complex concepts such as causality analysis. We believe that the ability to programatically generate questions is fundamental to assessing and improving LLM's ability to understand and reason about time series data. \TimeSeriesExam ~is available on \url{https://huggingface.co/datasets/AutonLab/TimeSeriesExam}
\end{abstract}


\section{Introduction}

There has been a lot of interest in exploring connections between Large Language Models (LLMs) and time series modeling. Most recent time series foundation models, for instance, leverage LLM architectures and are pre-trained on copious amounts of time series data \cite{ansari2024chronos, goswami2024moment, rasul2023lagllama, woo2024moirai, liutimer, garza2023timegpt}. Meanwhile, other studies have shown that when effectively ``reprogrammed'', LLMs can directly excel at various time series tasks including forecasting \cite{jin2023timellm, cao2023tempo, chang2023llm4ts}, anomaly detection, imputation, and classification \cite{zhou2023onefitsall}. But perhaps the most unexpected result was that LLMs can be zero-shot forecasters \cite{gruver2024llmtime}. These surprising results raise a fundamental question: \textit{do LLMs understand basic time series concepts?}

Existing attempts to answer this question have relied on domain-specific benchmarks~\cite{oh2024ecgqa, xing2021deepsqa} or synthetic datasets generated using LLMs themselves~\cite{mike2024llmstruggle}. However, performance on these benchmarks is not a perfect measure of an LLM's understanding of time series data. This is because they are confounded by the additional domain knowledge required to answer domain-specific questions (e.g., understanding of what the \emph{arrhythmia} is in the context of ECG data). Furthermore, synthetic time series generated using LLMs may not be entirely accurate, as their correctness depends on the very ability we aim to evaluate. Finally, these benchmarks are static, offering little to no control over the qualitative properties (e.g., difficulty) of their inquiries.

To bridge this gap, we introduce \TimeSeriesExam, a scalable and configurable multiple-choice question exam to assess LLMs across five core time series understanding categories. \TimeSeriesExam~comprises of over 700 questions, procedurally generated using carefully designed templates, and refined using Item Response Theory (IRT)~\cite{embretson2013itemrt, guinet2024autorag} to ensure each question has an appropriate level of difficulty and effectively differentiates candidate LLMs with varying abilities. 

\begin{wrapfigure}[22]{r}{0.50\textwidth}
\centering
\includegraphics[width=0.5\textwidth]{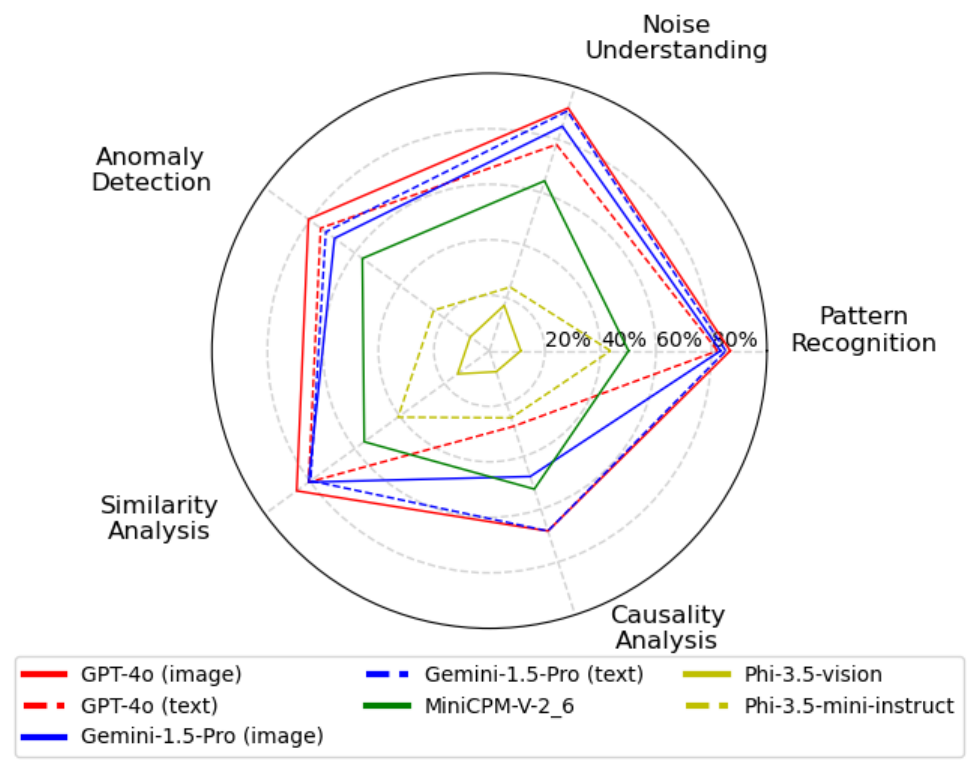}
\caption{Accuracy of latest LLMs on the \TimeSeriesExam. Closed-source LLMs outperform open-source ones in simple understanding tasks, but most models struggle with complex reasoning tasks.}
\label{figure: final round performance}
\end{wrapfigure}

We evaluate 7 state-of-the-art open and closed-source LLMs on the \TimeSeriesExam ~and provide the first comprehensive evaluation of their ability to understand basic time series concepts. Our findings reveal a decisive capability gap between closed and open-source models in understanding simple time series concepts. For brevity, we defer a detailed discussion of related work to App.~\ref{app:related_works}.

\section{\TimeSeriesExam: A Scalable and Configurable Time Series Exam  }

\begin{table}[bt!]
\resizebox{\columnwidth}{!}{%
\begin{tabular}{@{}r|   cc@{}}
\toprule
\textbf{Category}                            & \textbf{Subcategory}                              & \textbf{Example question}                                                            \\ \midrule
\multirow{9}{*}{Pattern Recognition} &
  Trend &
  What is the most likely linear trend coefficient of the given time series? \\ \cmidrule(l){2-3} 
 &
  Cyclic &
  \begin{tabular}[c]{@{}c@{}}The given time series has sine wave pattern. \\ How does its amplitude change from the beginning to the end?\end{tabular} \\ \cmidrule(l){2-3} 
                                     & Stationarity                   & Is the given time series likely to be stationary after removing the cycle component? \\ \cmidrule(l){2-3} 
                                     & Regime Switching               
                                     & Based on the given time series, how many different regimes are there?                \\ \cmidrule(l){2-3} 
                                     & Statistical properties 
                                     & Is the mean stable over time in the given time series?                               \\ \cmidrule(l){2-3} 
                                     & Random processes 
                                     & Does the following time series exhibit a mean reversion property?                    \\ \midrule
\multirow{5}{*}{Noise Understanding} & White Noise                 & Is the given time series a white noise process?                                      \\ \cmidrule(l){2-3} 
                                     & Random Walk                  & Is the given time series likely to be a random walk process?                         \\ \cmidrule(l){2-3} 
 &
  Signal / Noise Ratio &
  \begin{tabular}[c]{@{}c@{}}You are given two time series with the same underlying pattern but different noise level. \\ Which time series has higher magnitude of noise?\end{tabular} \\ \midrule
Anomaly Detection &
   &
  \begin{tabular}[c]{@{}c@{}}The following time series has two types of anomalies appearing at different time points. \\ What are the likely types of these anomalies?\end{tabular} \\ \midrule
\multirow{3}{*}{Comparative Analysis} & Shape                                    & Despite the noise, do the given two time series have similar patterns?              \\ \cmidrule(l){2-3} 
 &
  Distributional &
  \begin{tabular}[c]{@{}c@{}}You are given two time series which are generated using a random walk. \\ Are they likely to have the same variance?\end{tabular} \\ \midrule
Causality Analysis                   & Granger Causality                        & Is there Granger causality between the two time series?                                  \\ \bottomrule
\end{tabular}%
}
\caption{Example template questions for different reasoning tasks. Each subcategory covers a specific aspect of time series understanding, guiding the model to reason about comparative, anomalies, and causal relationships. }
\label{tab:taxonomy}
\end{table}

\paragraph{Composition.} The \TimeSeriesExam~ systematically assesses whether LLMs \textbf{understand} basic time series patterns such as trends and seasonality (\textit{pattern recognition}), the concept of noise and other time series concepts in the presence of noise (\textit{noise understanding}). It also evaluates LLMs on three different \textbf{reasoning} tasks: identifying abrupt deviation from ``normal" behavior~\cite{goswami2022unsupervisedmono} (\textit{anomaly detection}), comparing and contrasting statistical properties of 2 time series (\textit{comparative reasoning}), reasoning about causality, specifically Granger Causality~\cite{granger1969granger} (\textit{causality}). We expect these five categories will present increasing levels of difficulty for LLMs, particularly the reasoning tasks,  which typically involve multiple time series and require a grasp of basic time series concepts. As shown in Tab.~\ref{tab:taxonomy}, each category is further divided into sub-categories that represent more specific concepts within the broader category.

\begin{wrapfigure}[20]{r}{0.65\textwidth}
\centering
\includegraphics[scale=0.45]{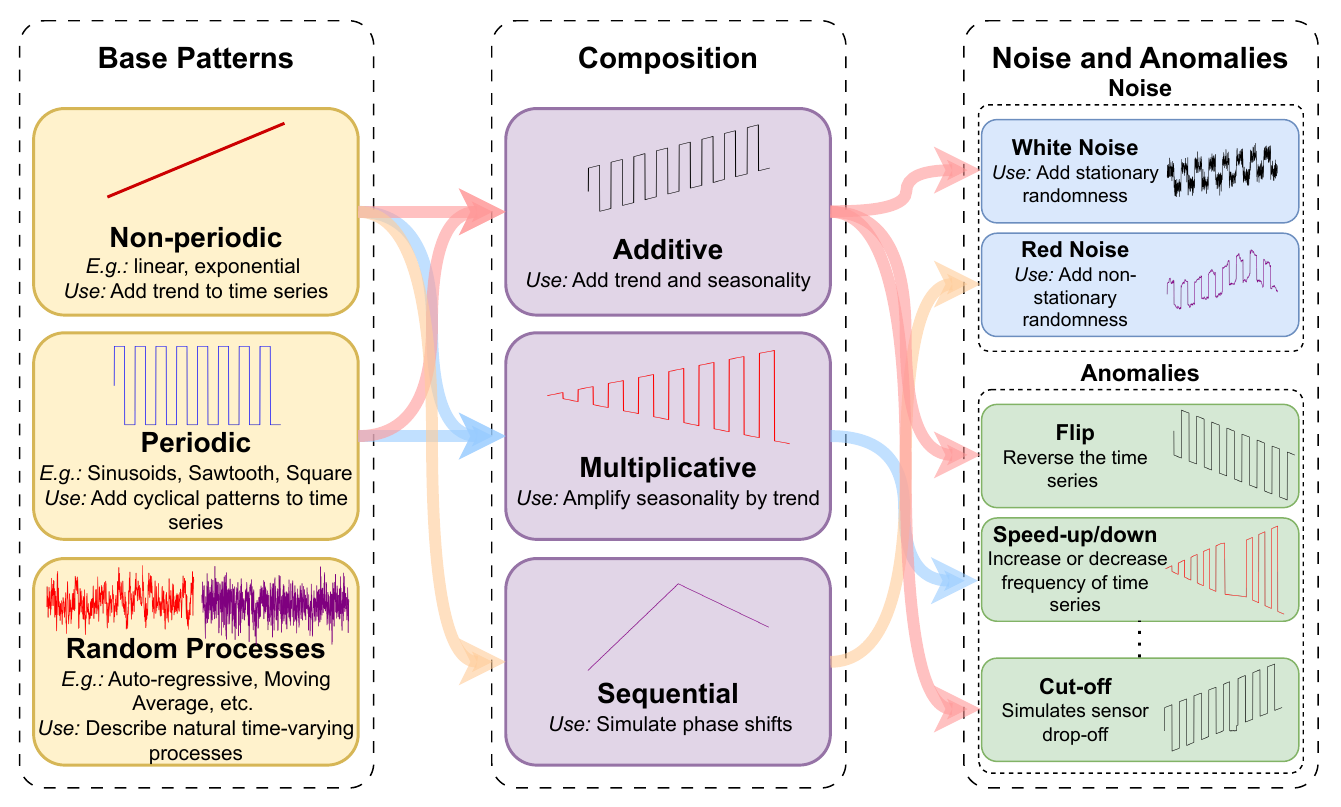}
\caption{Time Series Curation Pipeline: The composition model generates controlled synthetic time series step-by-step. The pipeline enables diversity by combining different components to create numerous synthetic time series with varying properties.}
\label{figure: curation pipeline}
\end{wrapfigure}

\begin{wrapfigure}[20]{r}{0.53\textwidth}
\centering
\includegraphics[scale=0.5]{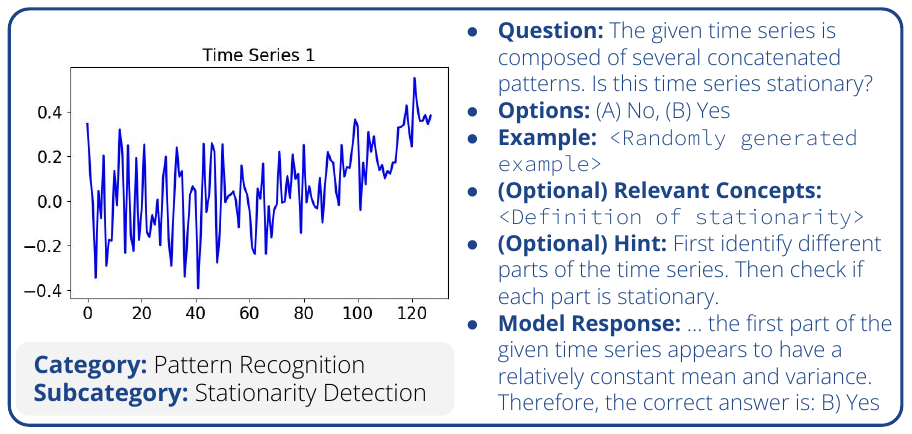}
\caption{Each template evaluates a specific category, and includes a question, list of options, example question and answer pair for in-context learning, and optionally a hint and descriptions of complicated technical terms. Here, \texttt{GPT-4o} 
showcases its ability to transfer visual understanding and time series concepts into effective reasoning.}
\label{figure: template}
\end{wrapfigure}

\paragraph{Question Templates.} The \TimeSeriesExam~ comprises over 100 unique templates, carefully curated in collaboration with time series experts and cross-verified for accuracy, that can be used to generate any number of random questions. Each template (Fig.~\ref{figure: template}) evaluates a specific (sub-)category (e.g.,~\textit{pattern recognition}), and comprises of a question (e.g., \texttt{``Is this time series stationary?"}), a list of options (e.g., \texttt{``(A) Yes, (B) No"}), and an example question and answer pair for in-context learning. Each template comes with a \textit{hint} which breaks down complex questions into simpler steps and textual descriptions of complicated technical terms. By incorporating these relevant concepts, we can isolate an LLM's ability to understand time series concepts (e.g., whether the mean and variance remain constant) from its understanding of complex technical jargon (e.g., stationarity). Each option (e.g. \texttt{``(A) Yes"}) is linked to a synthetic time series generator (Fig.~\ref{figure: curation pipeline}) that produces a random time series as if the current option were true (e.g., a random stationary time series). This allows us to generate random but accurate time series at scale.


\paragraph{Generating Questions.}
We generate different questions from the same template by systematically varying the correct option and producing synthetic time series conditioned on the template and the correct option pair. Our simple and scalable approach, illustrated in Fig.~\ref{figure: curation pipeline}, involves sampling a small number of base patterns from a predefined pool and combining them using a composition function. Base patterns can be periodic (e.g., sine function), non-periodic (e.g., linear increasing function), or random time-varying processes (e.g., AR process). Depending on the template's nature, the final step adds additive noise or anomalies using the anomaly injection process described in~\cite{goswami2022unsupervisedmono}.

\paragraph{Improving Questions Iteratively.} We use Item Response Theory (IRT)~\cite{lord2008statisticalirt} to achieve finer grained control over the quality of randomly generated questions included in the \TimeSeriesExam. IRT is a statistical framework that models the relationship between an individual's (or LLM's) latent trait (e.g., knowledge, ability) and their responses to a set of items (e.g., questions on a test). It is a valuable tool in exam development as it helps to identify weak exam items, ensures consistent scoring across different versions of the exam, and also allows  tailoring the testing experience to the LLM's abilities. 

Our primary objective is to design a \TimeSeriesExam~ where each question can maximally distinguish the abilities of candidate LLMs. We use the two-parameter logistic (2PL) model for this. Formally, for LLM $j$ with ability $\theta_j$, and question $i$ with difficulty $b_i$, discrimination ability $a_i$, the 2PL model defines the probability of a correct response as:
$\label{equation: irt 2pl}
    \mathbb{P}(r_{ij} = 1 | a_i, b_i, \theta_j) = 1 / (1+e^{-a_i(\theta_j - b_i)})$

Each \TimeSeriesExam~ typically undergoes 1--3 rounds of iterative refinement. In each round, all candidate models take the exam. Based on their responses, we fit the parameters of Equation~\ref{equation: irt 2pl} using maximum likelihood estimation (MLE). Then, we drop $X\%$ of samples with the lowest sum of difficulty and discrimination ability. Finally, we randomly re-generate questions from the dropped templates. This iterative process is detailed in Algorithm~\ref{alg:irt_resampling} and the hyper-parameters of the fitting process are provided in App.~\ref{appendix: irt fitting}.


\section{Experiments and Results}

\paragraph{Experimental Setup.} We evaluate LLMs on \TimeSeriesExam~ using two different setups for feeding time series into language models: (1) Image, where time series are plotted and input as images; and (2) Text, where time series values are truncated to one decimal place and separated by commas. We evaluate two proprietary models, \texttt{GPT-4o}~\cite{achiam2023gpt} and \texttt{Gemini}~\cite{reid2024gemini}. For open source models, we chose \texttt{Phi3.5}~\cite{abdin2024phi}, and \texttt{MiniCPM}~\cite{yao2024minicpm}. We selected these models to study: (1) the impact of model size on time series understanding, and (2) the effect of time series tokenization on model performance. Previous studies, such as \texttt{LLMTime}~\cite{gruver2024llmtime}, have explored tokenizing time series as text. We chose not to scale the time series, as their magnitudes are intentionally small to conserve tokens, and scaling could distort the shape, which is critical for many of the questions. The evaluations are done with one shot setting. The experiment details are provided in App.~\ref{appendix: experiment detail}. 

\paragraph{Impact of Model Size and Time Series Tokenization on Reasoning}

Fig.~\ref{figure: final round performance} reveals two key findings: first, closed-source models like \texttt{GPT-4o} and \texttt{Gemini} outperform open-source models in basic tasks such as pattern recognition but all models struggle with more complex tasks like causality analysis. This performance gap can likely be attributed to differences in pretraining data and model size, as seen in the comparison between \texttt{MiniCPM} (7B parameters) and \texttt{Phi-3.5} (3.8B parameters). Second, tokenizing time series data as images generally produces better results than textual tokenization. We offered an example response from GPT-4o using both tokenization method in Fig.~\ref{figure: text vs image}. Our qualitative analysis suggests that this outcome is likely due to the text-based approach causing models to focus excessively on details. This finding suggests that tokenization strategy is a critical factor in advancing reasoning capabilities, and they point to the potential for multimodal models that integrate time series and text data for more robust interactive reasoning.

\begin{wraptable}[15]{r}{0.39\textwidth}
\begin{center}
\scalebox{0.75}{
\begin{tabular}{r|c}
\toprule
\textbf{Question \& Guidance} & \textbf{Accuracy} \\ \midrule
Query only                         & 79.16            \\ 
Query + Hint                     & \textbf{79.29}   \\ 
Query + Relevant Concepts        & 75.88            \\ 
Query + Hint + Relevant Concepts & 77.71            \\ \bottomrule
\end{tabular}%
}
\caption{\texttt{Gemini-1.5-Pro} Performance with Different Dataset Components: The results indicate that adding relevant concepts, alongside hints, can sometimes decrease the model's performance. This may occur because the introduced concepts either contradict the model's internal knowledge or misguide its reasoning, leading to an incorrect chain of thought.}
\label{tab: accu with different element}
\end{center}
\end{wraptable}

\paragraph{Iterative Improved Benchmark} We observed an increasing trend for the sample average discrimination parameter($a_i$ in Equation~\ref{equation: irt 2pl}), which reflects the model's capacity to distinguish between individuals with varying ability levels. As the discrimination parameter increases, the model's ability to differentiate improves. The figure for the same can be found in App.~\ref{appendix: sample average discrimination parameter}.

\paragraph{Effect of Guidance on Model Reasoning} Tab.~\ref{tab: accu with different element} shows the performance of \texttt{Gemini-1.5-Pro} with various dataset elements provided. Question hints, which offer the first step in the reasoning process, improved the model’s performance, suggesting that sometimes model could struggle to form a coherent, logical sequence to reach the correct answer. Interestingly, providing relevant concepts hindered performance, possibly due to confusion or discrepancies with the model’s pretraining data.

\section{Conclusion}

In this work, we introduced a controlled, salable time series benchmark. We demonstrated that proprietary models like \texttt{Gemini} and \texttt{GPT-4} achieved non-trivial performance when time series were provided as both images and text. However, all models still struggle with complex reasoning tasks requiring multiple time series and multi-step inference. This highlights some key future directions: 

\paragraph{Developing a Benchmark for Practical and Complex Reasoning Tasks} While this work emphasizes reasoning based on time series understanding of patterns, future benchmarks should address more advanced tasks including complex causality analysis and context-driven forecasting.

\paragraph{Designing a More Rigorous Exam} We refer to our benchmark as an examination due to its structured components that scientifically assess a range of abilities. 
Future benchmarks should adopt more rigorous designs that query specific knowledge. Drawing from concepts such as knowledge tracing, we can introduce purposefully designed detractors to evaluate model performance better. 


\section{Acknowledgement}
This work was partially supported by the NSF (awards 2406231 and 2427948), and the US Army (W911NF-20-D0002).

\bibliographystyle{unsrt}
\bibliography{reference}



\appendix
\newpage
\section{Related Works}
\label{app:related_works}
\paragraph{Time Series Reasoning Benchmark.}
There exist a few benchmarks for time series reasoning, including a recent work that categorized time series reasoning into three primary tasks: \textit{context-aided forecasting}, \textit{question answering}, and \textit{etiological reasoning}~\cite{mike2024llmstruggle}. However, the work was limited by the fact that all samples were generated using \texttt{GPT}, where scientific design and correct time series correspondence are not guaranteed. Other efforts towards building time series reasoning benchmarks were exclusively for domain-specific tasks, such as those defined in \texttt{ECGQA}~\cite{oh2024ecgqa}. Thus, no existing benchmark currently evaluates whether LLMs possess an innate understanding of time series concepts and can transfer that understanding into structured reasoning. Our work bridges this gap by proposing a synthetically generated controlled dataset for this evaluation.

\paragraph{Synthetic Time Series Generation.}
The generation of synthetic time series with controlled behaviors, such as trends and cyclic patterns, is fundamental for constructing accurate reasoning benchmarks. A common approach involves sampling from diverse random processes~\cite{ismail2020synthetictsbenchmark}, such as Autoregressive Processes, which offer variability but lack control over specific patterns like cyclic behavior. To address this, \cite{woo2022cost} proposed a decomposition-based method, generating desired patterns by incorporating cyclic components into an additive model on top of random processes. We build upon both works by having a more diverse set of random processes and patterns, incorporating not only additive composition methods but also multiplicative and other forms of composition.

\paragraph{Time Series Foundation Models.} With the advent of several time series foundation models~\cite{goswami2024moment,ansari2024chronos,rasul2023lagllama,cao2023tempo, woo2024moirai, das2023googlefm, liutimer, garza2023timegpt, talukder2024totemtokenizedtimeseries} in recent months, there has been a paradigm-shifting change in the ability of models to perform time series analysis (TSA) tasks like forecasting, classification, imputation, and anomaly detection, including in zero-shot settings. These models apply language model architectures pre-trained on time series data to time series analysis tasks, achieving state-of-the-art performance compared to architectures built exclusively for time series analysis like \texttt{Informer}~\cite{zhou2021informer}. While their performance benefits for these tasks are noticeable, understanding of their reasoning ability and capability of identifying the nuances of a time series are yet to be evaluated. This is exacerbated by the issue that their outputs are typically time series, which is difficult for users to understand, rather than a response explaining the time series and characteristics that lead to the given output. LLMs' ability to reason is important here, since they generate responses that users can easily understand.

\section{Dataset Details}
\label{appendix: time series components}
Tab.~\ref{tab: category break down} presents the meta information for each category, while Tab.~\ref{tab: time series object} outlines the components of time series synthesis. The dataset includes 11 unique base patterns, 3 composition methods, 10 transformations, and 2 paired time series creation methods. These combinations produce a diverse set of time series with controlled features such as trend and seasonality.

\begin{table}[h]
\resizebox{\columnwidth}{!}{%
\begin{tabular}{rccccc}
\toprule
\textbf{Meta-information Type} & \textbf{Anomaly Detection} & \textbf{Similarity Analysis} & \textbf{Noise Understanding} & \textbf{Pattern Recognition} & \textbf{Causality Analysis} \\
\midrule
\# Questions & 129 & 113 & 87 & 371 & 63 \\
\%age questions with 2 time series & 31.01 & 100 & 14.94 & 3.77 & 100 \\
\bottomrule
\end{tabular}
}
\caption{TimeSeriesExam meta-information breakdown for each category. Each question is associated with a time series of length 128 time steps, and an example time series of length 64 time steps.}
\label{tab: category break down}
\end{table}

\begin{table}
\resizebox{\columnwidth}{!}{
\begin{tabular}{rl}
\toprule
\textbf{Object Name}                   & \textbf{Description}     \\ \hline
\multicolumn{2}{c}{\textbf{Base Time Series Objects}}    \\
Linear Trend                  & Create a linear line following $y=at$               \\ 
Exponential Trend             & Create an exponential line following $y=e^{at}$               \\ 
Log Trend                     & Create a log trend following $y=\log{t}$               \\ 
Constant                      & Create a constant line following $y=a$               \\
Gaussian White Noise          & Create an independent draw from Gaussian each time step               \\ 
Red Noise (Random Walk)       & Create a random walk where noise is drawn from normal, $y_t = y_{t-1} + \epsilon_t$               \\ 
SineWave                      & Generate a smooth, periodic oscillation following the function $y=A\sin(t)$              \\ 
Sawtooth Wave                 & Create a periodic waveform with a linear rise and abrupt fall               \\ 
SquareWave                    & Generate a periodic waveform that alternates between two levels, with equal durations               \\ 
Moving Average Process        & Create a moving average MA(q) random process, $y_t = \sum_{i=1}^{q}\epsilon_{t-i}\alpha_i + \epsilon_t$              \\ 
Autoregressive Process        & Create an autoregressive AR(p) random process, $y_t = \sum_{i=1}^{p}y_{t-i}\alpha_i + \epsilon_t$              \\ \hline
\multicolumn{2}{c}{\textbf{Composition Method}}          \\ 
Additive                      & Add each time series object at each time stamp             \\ 
Multiplicative                & Multiply each time series object at each time stamp              \\ 
Concatenate                   & Concatenate each time series object along time series horizon               \\ \hline
\multicolumn{2}{c}{\textbf{Transformation}}              \\ 
Anomaly Creation              & Following the nine synthetic transformations defined in~\cite{goswami2022unsupervisedmono}               \\
Sign Flip                     & $y_t = -y_t$               \\ \hline
\multicolumn{2}{c}{\textbf{Paired Time Series Creation}} \\
Lagged Pair Creation          & Create a lagged time series from a baseline time series              \\ 
Granger Pair Creation         & Create a granger caused time series from a baseline time series               \\
\bottomrule
\end{tabular}
}
\caption{Time Series in the \TimeSeriesExam are created from a combination of diverse baseline Time Series Objects. The baseline objects cover linear/non-linear signals and cyclic patterns. For trend variables, $t$ is a sequence of integers that represents time. $\epsilon_t$ is a standard normal random variable. For MA(q) and AR(p) processes, $\alpha_i$ represents its parameter.}
\label{tab: time series object}
\end{table}

\subsection{Iterative Refinement Algorithm}
\begin{algorithm}[H]
\caption{Iterative Dataset Refinement with IRT and Resampling}
\label{alg:irt_resampling}
\begin{algorithmic}[1]
\REQUIRE num\_iterations = 3, drop\_percentage = 0.2, initial dataset $D_0$
\STATE $D \gets D_0$
\FOR{$\text{iteration} = 1$ to num\_iterations}
    \STATE \textbf{Evaluate} each candidate $i$ on $D$, and obtain the response set $R = \{r_{ij} \mid r_{ij} = 1 \text{ if candidate } i \text{ correctly answers question } j\}$

    \STATE \textbf{Fit} the IRT model to obtain the discrimination parameters $\mathbf{A} = \{a_j \mid j \in \text{Questions}\}$ and difficiulty parameter $\mathbf{B} = \{b_j \mid j \in \text{Questions}\}$

    \STATE \textbf{Normalize} set $\mathbf{A}$ and $\mathbf{B}$ between 0 and 1, and calculate score $\mathbf{S} = \{b_j + a_j \mid j \in \text{Questions}\}$

    \STATE \textbf{Find} $\mathbf{S'}$ which is the score for samples that are answered correctly by the best model in the round

    \STATE \textbf{Find} the index set $I = \{j \mid a_j < \text{Quantile}(\mathbf{S'}, \text{drop\_percentage})\}$, where $a_j$ is less than the $\text{drop\_percentage}$ quantile of $\mathbf{A}$

    \FOR{each $j \in I$}
        \STATE \textbf{Resample} a new question $q'$ from the same category as question $j$
        \STATE Set $D[j] \gets q'$
    \ENDFOR

\ENDFOR
\RETURN $D$
\end{algorithmic}
\end{algorithm}

\section{Experiment Details}
\label{appendix: experiment detail}
\subsection{Generation Configuration}
We set the maximum token length to 1024 and a temperature of 0.0 for generation. For models that support seed control, we use a seed value of 42; otherwise, seed control is unavailable in some proprietary models \footnote{The evaluation code is available at \url{https://anonymous.4open.science/r/TimeSeriesExam-8387}}

\subsection{IRT Model Parameters}
\label{appendix: irt fitting}
The IRT models are fitted using library $\textit{py-irt}$~\cite{lalor2023py-irt-lib}. The parameters are epochs=2000, lr=0.1, lrdecay=0.9999, dropout=0.5, hidden=100

\subsection{Average Sample Discrimination Parameter over Rounds}
\label{appendix: sample average discrimination parameter}
\begin{figure}[h!]
\centering
\includegraphics[scale=0.30]{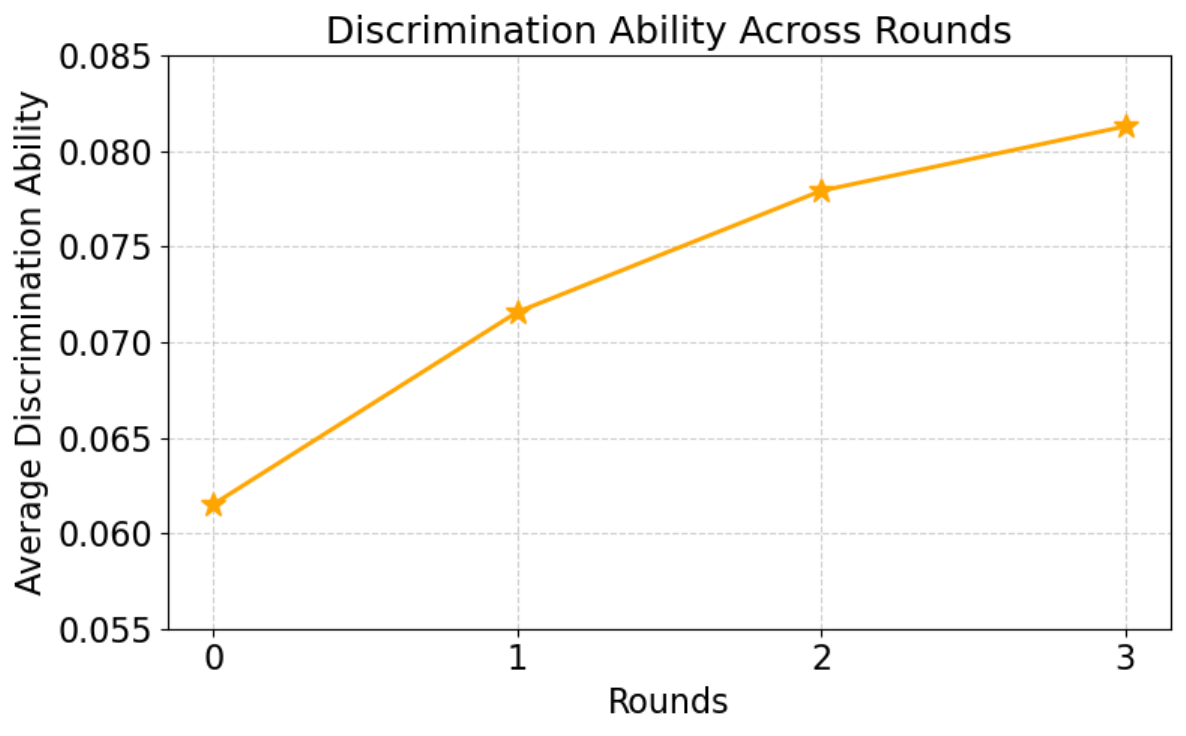}
\caption{The sample average discrimination parameter across rounds shows an upward trend, indicating an improved ability of the questions to differentiate candidates with varying levels of ability.}
\label{figure: discrimination over rounds}
\end{figure}

\subsection{Dropped dataset distribution per round}

\begin{figure*}[h!]
\centering
\includegraphics[scale=0.60]{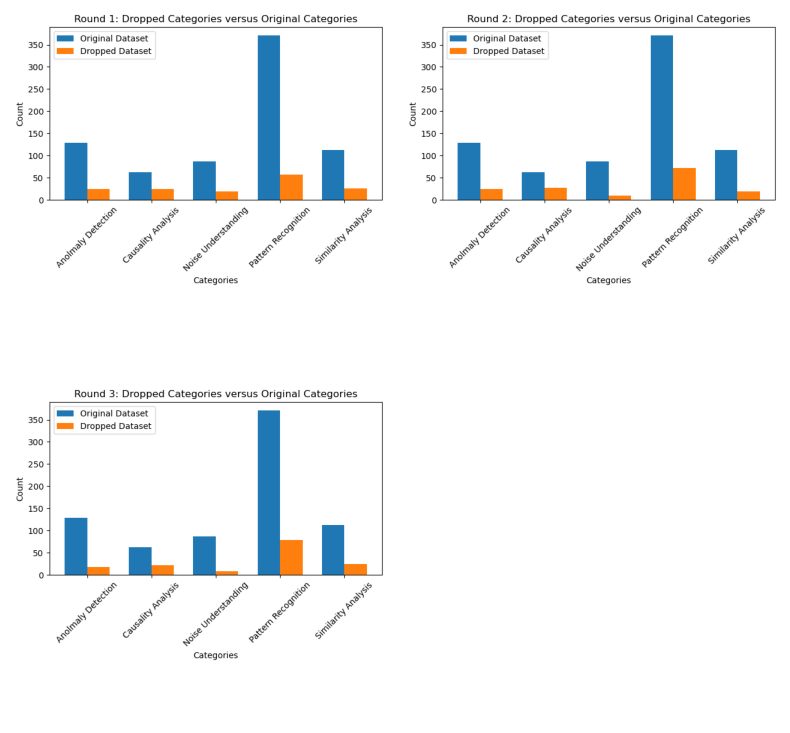}
\caption{Dropped Dataset Distribution per round. Dropped category distribution per round generally mirrors the overall category distribution.}
\label{figure: drop data set distribution}
\end{figure*}

We can observe in Fig.~\ref{figure: drop data set distribution} that the proportion of dropped questions for each category is approximately uniform.

\subsection{Inference Cost}
\begin{table}[]
\centering
\resizebox{0.6\linewidth}{!}{
\begin{tabular}{c|c|c}
\toprule
\textbf{Model}     & \textbf{Image Tokenization Cost (\$)} & \textbf{Text Tokenization Cost (\$)} \\ \hline
\texttt{GPT-4o}    & 0.009703598             & 0.008769533            \\ 
\texttt{Gemini}    & 0.006792518             & 0.006138673            \\ \bottomrule
\end{tabular}%
}
\vspace{10pt}
\caption{Inference cost per sample (in USD) based on tokenization methods. Costs for both image and text tokenization are similar and cost-effective. Token prices were sourced from official model documentation.}
\label{tab: cost}
\end{table}

We report inference cost per sample for proprietary models in~\ref{tab: cost}. The average token size per sample for image tokenization is 1940.72, the average token size per sample for text tokenization is 1753.91. Number of tokens are calculated based on \texttt{GPT4} tokenizer.

\clearpage
\subsection{Model Accuracy per Round with Category Break down}

\begin{table}[h]
\resizebox{\columnwidth}{!}{%
\begin{tabular}{rcccccc}
\toprule
\textbf{Model} & \textbf{Pattern Recognition} & \textbf{Noise Understanding} & \textbf{Anomaly Detection} & \textbf{Similarity Analysis} & \textbf{Causality Analysis} \\ \midrule
\multicolumn{6}{c}{\textbf{Round 0}}                               \\ \midrule
\texttt{Phi-3.5 (Image)}         & 0.20 & 0.34 & 0.11 & 0.19 & 0.22 \\ 
\texttt{Phi-3.5 (Text)} & 0.46 & 0.25 & 0.27 & 0.45 & 0.52 \\ 
\texttt{Gemini-1.5-Pro (Image)}            & 0.81 & 0.82 & 0.64 & 0.78 & 0.59 \\ 
\texttt{Gemini-1.5-Pro (Text)}             & 0.75 & 0.85 & 0.62 & 0.65 & 0.52 \\ 
\texttt{GPT-4o (Image)}                    & \textbf{0.88} & \textbf{0.92} & \textbf{0.78} & \textbf{0.87} & \textbf{0.86} \\ 
\texttt{GPT-4o (Text)}                   & 0.74 & 0.82 & 0.65 & 0.71 & 0.19 \\ 
\texttt{MiniCPM (Text)}          & 0.53 & 0.67 & 0.56 & 0.55 & 0.54 \\ 
\midrule
\multicolumn{6}{c}{\textbf{Round 1}}                               \\ \midrule
\texttt{Phi-3.5 (Image)}         & 0.16 & 0.25 & 0.13 & 0.16 & 0.14 \\ 
\texttt{Phi-3.5 (Text)} & 0.47 & 0.26 & 0.28 & 0.40 & 0.49 \\ 

\texttt{Gemini-1.5-Pro (Text)}          & 0.78 & 0.89 & 0.67 & 0.77 & 0.60 \\ 
\texttt{GPT-4o (Text)}                   & 0.77 & 0.84 & 0.72 & 0.77 & 0.21 \\ 
\texttt{GPT-4o (Image)}                 & \textbf{0.84} & \textbf{0.91} & \textbf{0.76} & \textbf{0.87} & \textbf{0.79} \\ 
\texttt{MiniCPM (Text)}             & 0.53 & 0.63 & 0.59 & 0.54 & 0.48 \\ 
\midrule
\multicolumn{6}{c}{\textbf{Round 2}}                               \\ \midrule
\texttt{Phi-3.5 (Image)}         & 0.14 & 0.22 & 0.09 & 0.15 & 0.10 \\ 
\texttt{Phi-3.5 (Text)} & 0.47 & 0.26 & 0.24 & 0.42 & 0.38 \\ 
\texttt{Gemini-1.5-Pro (Image)}         & 0.89 & 0.86 & \textbf{0.74} & \textbf{0.87} & 0.67 \\ 
\texttt{Gemini-1.5-Pro (Text)}          & \textbf{0.83} & \textbf{0.91} & 0.72 & 0.82 & 0.67 \\
\texttt{GPT-4o (Image)}                   & 0.81 & 0.90  & 0.73 & 0.83 & \textbf{0.68} \\
\texttt{GPT-4o (Text)}                  & 0.80  & 0.83 & 0.73 & 0.81 & 0.25 \\  
\texttt{MiniCPM (Text)}          & 0.51 & 0.64 & 0.57 & 0.53 & 0.54 \\ 
\midrule
\multicolumn{6}{c}{\textbf{Round 3}}                               \\ \midrule
\texttt{Phi-3.5 (Image)}         & 0.11 & 0.17 & 0.09 & 0.14 & 0.08 \\ 
\texttt{Phi-3.5 (Text)} & 0.44 & 0.24 & 0.25 & 0.41 & 0.25 \\ 
\texttt{Gemini-1.5-Pro (Image)}         & 0.83 & 0.85 & 0.69 & 0.81 & 0.48 \\ 
\texttt{Gemini-1.5-Pro (Text)}         & 0.85 & 0.91 & 0.73 & 0.80 & \textbf{0.68} \\ 
\texttt{GPT-4o (Image)}                & \textbf{0.87} & \textbf{0.92} & \textbf{0.81} & \textbf{0.86} & \textbf{0.68} \\ 
\texttt{GPT-4o (Text)}                  & 0.82 & 0.78 & 0.75 & 0.81 & 0.29 \\ 
\texttt{MiniCPM (Text)}            & 0.50 & 0.64 & 0.57 & 0.56 & 0.52 \\ 
\bottomrule
\end{tabular}%
}
\caption{Breakdown of model accuracy across multiple rounds of item response theory. }
\end{table}

\newpage
\subsection{Case Study: Model response under image tokenization and text tokenization}

\begin{figure*}[h!]
\label{figure: text vs image}
\centering
\includegraphics[width=\linewidth]{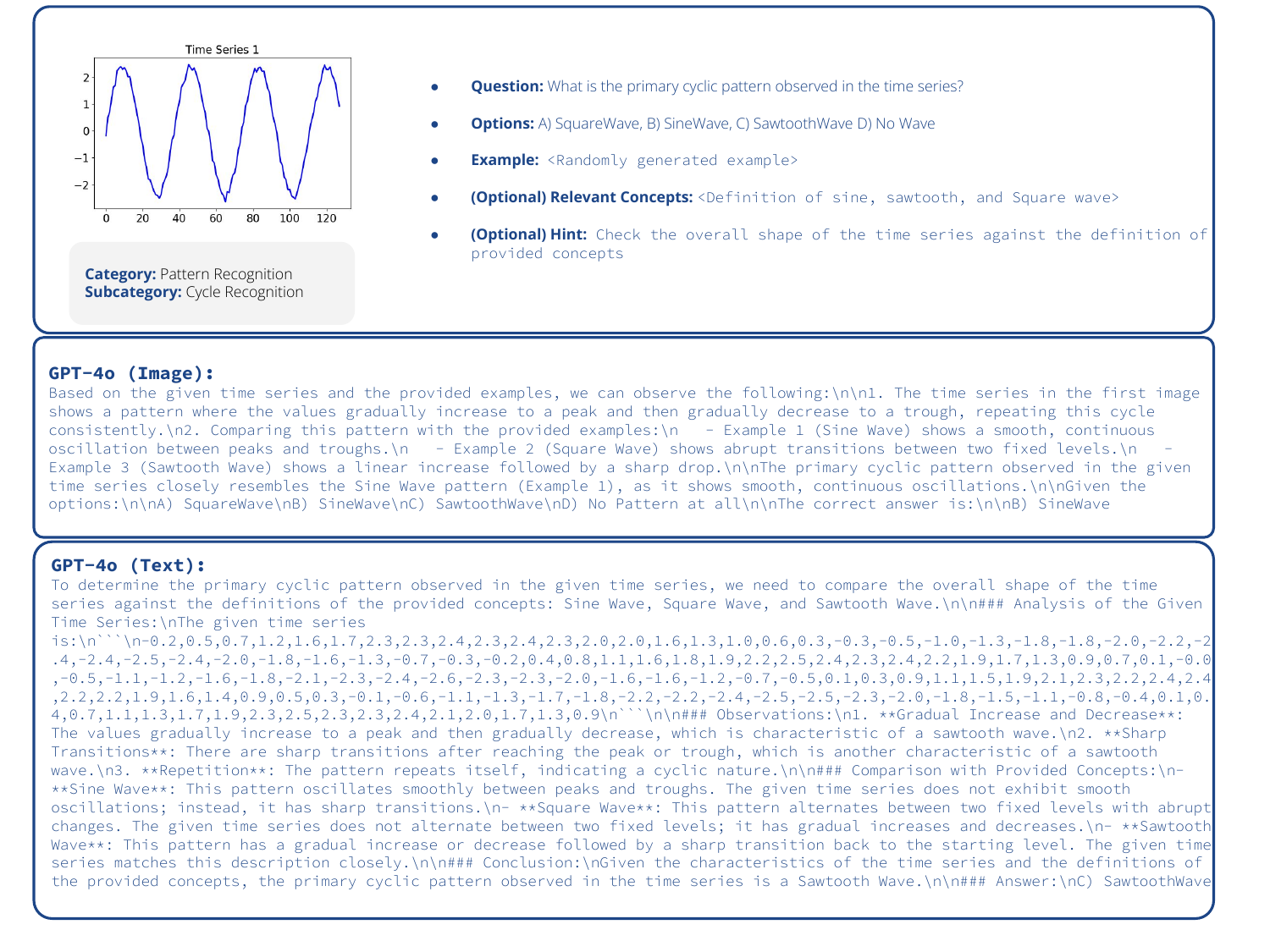}
\caption{The image-based model provided the correct answer, while the text-based model failed due to its inability to translate detailed numerical data into a specific trend shape.}
\label{figure: image vs text}
\end{figure*}
\end{document}